\documentclass[11pt]{article}

\usepackage[margin=1in]{geometry}
\usepackage[T1]{fontenc}
\usepackage{lmodern}
\usepackage{microtype}
\usepackage{amsmath,amssymb}
\usepackage{graphicx}
\usepackage{booktabs}
\usepackage{array}
\usepackage{multirow}
\usepackage{xcolor}
\usepackage{enumitem}
\usepackage[hidelinks]{hyperref}

\newcommand{\poweredTasks}{6}

\newcommand{\poweredEpisodes}{576}

\title{\textbf{When the Governor Becomes the Disturbance:}\\[0.3em]
Control-Generated Disturbance and Cost-Aware Backoff in Governed Tool-Using Agents}
\author{Veronique Ziegler\\Independent Researcher, Critical Attention Systems}
\date{\today}

\begin{document}
\maketitle

\begin{abstract}
Supervisory governors can interfere with the tool-using agents they regulate.
We study this possibility in a controlled file-recovery environment where
increases in regulatory intensity trigger experimentally imposed tool failures.
A cost-blind governor can turn these failures into persistent blocking that
prevents task completion. We compare this governor with a backoff rule that
reduces intervention probability using a moving average of known induced
events. On a hand-coded stochastic-policy agent, the failure pattern appears
under both result replacement and execution of corrupted tool arguments. For
the persistent policy, adaptive backoff improves completion relative to a fixed
weak governor with approximately matched intervention frequency. A Gemini~2.5
Flash experiment comprising $\poweredEpisodes$ episodes across
$\poweredTasks$ tasks also shows reduced blocking and improved completion
under backoff; among the tested settings, intermediate backoff strength
achieves the highest observed aggregate success. These results identify an
interaction between intervention cost and persistent action blocking, together
with a possible mitigation. The cost mechanisms are imposed and their induced
events are directly observable to the backoff rule; applicability beyond this
controlled environment remains an empirical question.
\end{abstract}

\section{Introduction}

Autonomous tool-using agents can be supervised by a
\emph{governor}: a monitor that inspects the agent's trajectory, emits a
regulatory signal, and can warn or block proposed actions to keep the agent
from looping, retrying failed assumptions, or answering prematurely. Feedback
control provides a natural language for this supervisory
role~\cite{astrom2008feedback}. The benefit of intervention, however, depends
on both the problem it prevents and the disturbance it introduces.

If a corrective intervention degrades an otherwise valid action and the
resulting failure enters the governor's history, later regulation can preserve
the consequences of that failure. In a repeat-blocking governor, even a
transient failure can prevent the intended action from being tried again.
The agent is no longer merely being regulated; \emph{the governor has become a
disturbance source}. We investigate this interaction with an explicit,
experimentally imposed cost of tightening control.

This question is a natural extension of IRAM-$\Omega$-Q, the Integrated
Regulation and Attention Model with $\Omega$-regulation and a quantum-like
state representation. This computational framework models uncertainty
regulation in adaptive agents: a normalized complex amplitude vector
represents the agent's internal state, and feedback from an entropy-based
uncertainty signal adjusts an adaptive regulatory gain $\mu(t)$. Here,
$\Omega$ denotes the adaptive regulation component, while the quantum-like
formalism supplies a mathematical language for the internal
state~\cite{ziegler2026iramomegaq}. The framework and its extensions study how
the timing and demand of regulation shape an agent's control
burden~\cite{ziegler2026carryover,ziegler2026switching}.

The possibility that regulation generates disturbance is also a standard
concern in feedback control, where an over-eager regulator can inject the very
disturbance it is trying to correct~\cite{astrom2008feedback}. We pose it as a
hypothesis---that a governor's own tightening, when it carries a cost, can
initiate failure that its retained action history and repeat blocking then
sustain---and ask whether the interaction appears with two different
action-selection mechanisms: a symbolic tool-using agent supervised by a
finite-state governor, and a real large-language-model (LLM) agent. Sustained
blocking need not involve further increases in regulatory intensity. The two
agents share the same task environment and governor; this comparison tests
whether the effect depends on a hand-coded action policy.

Our results are demonstrated on two substrates: a hand-coded stochastic-policy
substrate, where matched large-$N$ seeding is cheap, and a run on a real
LLM:
\begin{enumerate}[leftmargin=1.4em]
\item cost-blind governance produces increasing \emph{block-storm
collapse}---a decline in task completion caused by repeated governor blocks
that consume the step budget and prevent recovery---as the induced-cost scale
$\alpha$ grows, with both agent realizations; two cost implementations are
tested on the stochastic substrate
(Sections~\ref{sec:amplification},~\ref{sec:gemini});
\item cost-aware backoff reduces applied regulatory intensity and can rescue
completion (Sections~\ref{sec:suppression},~\ref{sec:gemini}); and
\item on the stochastic substrate, adaptive backoff outperforms a fixed weak
governor with approximately matched intervention frequency
(Section~\ref{sec:baselines}).
\end{enumerate}
The real-LLM experiment demonstrates collapse, suppression, and rescue. It
compares three backoff strengths, with the largest observed success rate at an
intermediate setting.

\section{Background and scope}
\label{sec:background}

Within IRAM-$\Omega$-Q, the adaptive gain $\mu(t)$ is interpreted as
regulatory demand. The model compares regulation-first (RF) ordering, in which
control attenuates incoming disturbance, with disturbance-first (DF) ordering,
in which control responds after exposure~\cite{ziegler2026iramomegaq,ziegler2026carryover}.
A recurring theme in that
line, and in feedback control more generally~\cite{astrom2008feedback}, is that
the \emph{source} of a disturbance matters, not only its magnitude---and in
particular that a controller's own tightening can itself act as a disturbance.
We model that tightening cost, in the continuous formalism's terms, as an
induced disturbance
$\eta_{\mathrm{induced}} =
\alpha\max(0,\mu_{\mathrm{after}}-\mu_{\mathrm{before}})^2$ charged only when
the controller tightens, and take it as the mechanism whose discrete analog we
study.

Our hypothesis is that a governor's own intervention cost can interact with
its retained action history to sustain failure, and that feedback-conditioned
backoff can interrupt this process. We do \emph{not} claim to reproduce the
continuous formalism's numerics or architecture; the discrete induced-cost
mechanism of Section~\ref{sec:method} is a deliberately simple analog of
$\eta_{\mathrm{induced}}$, and the correspondence is structural, not
quantitative.

\paragraph{Related work.}
The costs of protection have been studied in several settings. Defense-trained
LLM agents can exhibit early failures that propagate through retry loops and
timeouts~\cite{li2026autonomytax}. Selective guardrail methods seek to preserve
utility by avoiding unnecessary always-on intervention~\cite{kim2026causalarmor}.
In physical control, safety filters can induce unwanted oscillations, motivating
methods that account for the temporal consequences of
corrections~\cite{pizarro2023multistep}. Our focus is a specific discrete
interaction: experimentally imposed intervention failures are retained by a
repeat-blocking governor, and adaptive backoff changes the agent's opportunity
to recover.

\section{The governed tool-use substrate}
\label{sec:substrate}

\paragraph{Agent, governor, task.}
An episode places an agent in a small file workspace and asks it to locate and
summarize a target file; the prompt embeds an incorrect assumption (a wrong
filename, folder, extension, or a confidently-named nonexistent path), so
recovery requires inspecting the workspace. Each step the agent proposes one
tool action (\texttt{READ\_FILE} or \texttt{LIST\_DIRECTORY}) or a final
answer. A finite-state \emph{governor} evaluates the trajectory and emits a
regulatory state, with states \textsc{Normal}, \textsc{Disturbed}, \textsc{Recovering},
\textsc{Stuck}, and \textsc{Released}. \textsc{Disturbed} follows a failed
tool call; \textsc{Recovering} follows repeated outcomes or successful
uncertainty reduction after failure; \textsc{Stuck} is triggered by repeated
executed outcomes; and \textsc{Released} follows recovery of the target
content. Blocked proposals are recorded as interventions, not executed tool
failures. The benchmark governor is supplied with the expected target path to
identify successful recovery.
The governor can reshape the agent's action distribution toward inspection
(reweighting) and can \emph{block} an action that repeats a failed assumption
or continues tool use after \textsc{Released}. Task success is a substring
match of the final answer against the known target answer.

\paragraph{Two agent realizations.}
The \textbf{stochastic policy}~\cite{sutton2018rl} samples each step from a governor-modulated
weighted categorical over candidate actions; behavioral presets give
archetypes (an \emph{impulsive} high-entropy sampler; a \emph{persistent}
archetype that clings to failed actions). Its action-selection rules are fixed
rather than learned, and actions are sampled pseudorandomly, making thousands
of seeded runs inexpensive. It tests the mechanism under controlled policy
variation, not learned behavior. The \textbf{real LLM agent} delegates action
selection to Gemini~2.5~Flash via a JSON tool-call prompt embedding the current
regulatory signal.

\paragraph{Regulation intensity.}
We give the governor's tightening a scalar via a monotone intensity over its
state,
\begin{equation}
\begin{gathered}
g(\textsc{Normal})=0,\quad
g(\textsc{Disturbed})=0.75,\quad
g(\textsc{Recovering})=0.85,\\
g(\textsc{Stuck})=0.95,\quad
g(\textsc{Released})=0.99,
\end{gathered}
\label{eq:intensity}
\end{equation}
equal to one minus the governor's retry-suppression multiplier. Higher $g$
means harder clamping; this is the discrete counterpart of the continuous gain.

\section{Method: control-generated disturbance}
\label{sec:method}

\paragraph{Induced cost.}
Only \emph{positive} tightening generates cost. Here $g(t)$ denotes the
\emph{applied} intensity at decision step $t$, including zero on a backed-off
step. With
$\Delta g^{+}(t)=\max(0,\,g(t)-g(t-1))$, an induced event is drawn at each step
with probability
\begin{equation}
p_{\mathrm{induced}}(t)=\min\!\big(1,\;\alpha\,\Delta g^{+}(t)^2\big),
\label{eq:pinduced}
\end{equation}
$\alpha\ge0$ the single swept scale. At $\alpha=0$ no cost is ever generated
and the agent reduces exactly to the ordinary governed agent (a regression
anchor). Because the escalation $\textsc{Normal}\to\textsc{Disturbed}$ is a
large jump ($\Delta g^{+}=0.75$), firing saturates near $\alpha\approx1.8$, so
the informative range is $\alpha\in[0,2]$.

\paragraph{Two cost channels.}
An induced event is realized through one of two mechanisms, applied to the
agent's intended action so that the governor's repeat logic engages
identically:
\begin{itemize}[leftmargin=1.4em,itemsep=2pt]
\item \textbf{Result injection}: the intended action's result is replaced by a
fabricated transient failure.
\item \textbf{Argument corruption}: a bad-path (corruption-suffixed) version of
the intended action is \emph{actually executed} by the real tool, which returns
its own genuine not-found error; the failure is attributed to the intended
action. Nothing is fabricated---a genuinely degraded call really failed.
\end{itemize}
The two channels implement the same governor-visible failure assumption in
different ways. Their agreement tests implementation robustness within this
constructed cost model; it does not establish how often ordinary interventions
produce such failures.

\paragraph{Persistent blocking after induced failure.}
An induced failure is recorded in history against the intended action. The
governor then treats a retry of that action as repetition of a failed
assumption and can block it, even if the underlying fault was transient. A
sequence of such blocked proposals consumes the step budget without executing
the action needed for recovery: a \emph{block storm}. This persistence does
not require regulatory intensity to increase on every blocked step. Under
always-on repeat blocking, the tool history can remain unchanged and intensity
can stay constant after the initiating failure. Cost-aware backoff provides
steps on which the blocked action can instead be executed.

\paragraph{Cost-aware backoff.}
The cost-aware governor maintains an exponentially weighted induced-failure
rate $r$. With a smoothing factor of $\lambda=0.3$, it updates this rate after
each executed tool call as
\[
r_{\text{new}}=
\begin{cases}
0.7r_{\text{old}}+0.3 & \text{if an induced failure occurred},\\
0.7r_{\text{old}} & \text{otherwise}.
\end{cases}
\]
The event label is provided by the experiment, so this is an oracle attribution
condition. Let
$r(t)$ denote the latest stored estimate at the start of decision step $t$.
The governor applies its regulation---reweighting \emph{and}
blocking---only with probability
\begin{equation}
p_{\mathrm{apply}}(t)=1-\min\!\big(1,\;k\,r(t)\big),
\label{eq:papply}
\end{equation}
$k>0$ the backoff sensitivity (swept over $\{1.5,3.0,6.0\}$). On a backed-off
step reweighting and blocking are disabled, so the applied $g=0$ and no new
induced event is scheduled. The LLM receives a \textsc{Normal} current signal
while retaining its history. The estimate $r$ decays after a non-induced tool
execution; blocked proposals do not update it. An event scheduled before a
blocked proposal remains pending until the next executed tool call, including
one on a backed-off step. Thus, backoff prevents new scheduling but does not
cancel a previously scheduled event. Returning from a backed-off step to a
positive applied intensity counts as fresh tightening in
Equation~\ref{eq:pinduced}, even if the underlying regulatory state has not
changed. This feedback-conditioned bypass can interrupt the block storm.
A \textbf{fixed weak governor} that applies regulation
at a constant probability, non-adaptively, serves as a control for whether
adaptivity itself matters (Section~\ref{sec:baselines}).

\section{Demonstrated results I: stochastic substrate}
\label{sec:stochastic}

All results in this section use $2000$ seeds per (task, condition) cell
across eight file-recovery tasks, regulation-first (RF) governance, and a 16-step
budget. The backoff-sensitivity sweep uses matched seed indices across modes;
the mechanism and baseline comparisons use condition-specific random streams.

\subsection{Cost-blind governance amplifies into collapse}
\label{sec:amplification}

Under the cost-blind governor, raising $\alpha$ collapses success while the
governor's intensity stays pinned, and this holds under both cost channels
(Table~\ref{tab:mechanisms}). For the persistent archetype, success falls
$0.88\to0.49\to0.18$ (result injection) and $0.88\to0.48\to0.19$ (argument
corruption) as $\alpha$ goes $0\to1\to2$, with $\bar g$ pinned near $0.71$
throughout. The near-identity across a fabricated-failure channel and a
genuinely-executed-corrupted-action channel is consistent with their shared treatment of
induced governor-visible failure.

\begin{table}[t]
\centering
\caption{Mechanism robustness (persistent archetype, $2000$ seeds/cell $\times$
$8$ tasks). Success and mean intensity $\bar g$ by cost channel, backoff, and
$\alpha$. Cost-blind amplification and cost-aware suppression are essentially
identical under both channels.}
\label{tab:mechanisms}
\begin{tabular}{llccc}
\toprule
 & & $\alpha{=}0$ & $\alpha{=}1$ & $\alpha{=}2$ \\
\midrule
\multirow{2}{*}{Result injection}
 & cost-blind (succ / $\bar g$)  & $0.88\,/\,0.72$ & $0.49\,/\,0.71$ & $0.18\,/\,0.70$ \\
 & cost-aware (succ / $\bar g$)  & $0.88\,/\,0.72$ & $0.65\,/\,0.50$ & $0.40\,/\,0.29$ \\
\midrule
\multirow{2}{*}{Argument corruption}
 & cost-blind (succ / $\bar g$)  & $0.88\,/\,0.72$ & $0.48\,/\,0.71$ & $0.19\,/\,0.70$ \\
 & cost-aware (succ / $\bar g$)  & $0.88\,/\,0.72$ & $0.64\,/\,0.50$ & $0.40\,/\,0.29$ \\
\bottomrule
\end{tabular}
\end{table}

\subsection{Cost-aware governance suppresses its own intensity}
\label{sec:suppression}

Under the cost-aware governor, $\bar g$ falls with $\alpha$ (Table~\ref{tab:mechanisms},
persistent: $0.72\to0.50\to0.29$) while the cost-blind $\bar g$ stays pinned---the
suppression branch, absent by construction under cost-blindness. Suppression
deepens monotonically with backoff sensitivity $k$ (impulsive $k{=}6$:
$0.42\to0.13$; persistent $k{=}6$: $0.72\to0.21$). Success rescue is
agent-dependent: the persistent archetype is rescued (Table~\ref{tab:mechanisms},
$\alpha{=}2$: cost-blind $0.18$ vs cost-aware $0.40$), whereas the impulsive
archetype---whose ungoverned success rate is low---is not. This is consistent
with backoff exposing a policy that performs poorly without regulation, but
the comparison does not isolate that explanation. These two presets suggest
that the agent's unregulated performance can condition the benefit of backoff.

\subsection{Comparison with approximately rate-matched fixed governance}
\label{sec:baselines}

Table~\ref{tab:baselines} places the induced-cost conditions on a full baseline
ladder at $\alpha=2$. The key control is the \emph{fixed weak governor}, which
intervenes at a constant probability approximately matched to cost-aware's realized
intervention rate (fixed $0.45$ versus adaptive $0.46$ for persistent in
Table~\ref{tab:baselines}). For the persistent archetype:
ordinary governance at $\alpha=0$ lifts success far above no governance
($0.88$ vs $0.31$); induced cost makes cost-blind governance actively harmful,
dropping success \emph{below} no governance ($0.19$); cost-aware backoff
restores it above the un-governed level ($0.40$); and the fixed
weak governor at approximately the same intervention rate reaches $0.30$.
Thus, approximately matching average application frequency does not reproduce the adaptive
result in this comparison. The two schedules can still differ in switching
pattern, mean intensity, and which actions are permitted; the comparison does
not establish superiority over every fixed or error-aware alternative. For
the impulsive archetype, backoff does not improve success over cost-blind
governance ($0.10$ versus $0.11$); both remain close to its ungoverned rate
($0.09$). Unlike the persistent case, cost-blind success does not fall below
the ungoverned rate in this comparison.

\begin{table}[t]
\centering
\caption{Baseline ladder at $\alpha=2$ (result injection, $2000$ seeds/cell
$\times$ $8$ tasks). The fixed weak governor approximately matches cost-aware's realized
intervention (apply) rate. For persistent, cost-aware ($0.40$) beats
rate-matched fixed ($0.30$), and cost-blind governance ($0.19$) is worse than no
governance ($0.31$).}
\label{tab:baselines}
\begin{tabular}{lcccc}
\toprule
 & \multicolumn{2}{c}{persistent} & \multicolumn{2}{c}{impulsive} \\
\cmidrule(lr){2-3}\cmidrule(lr){4-5}
condition & success & apply rate & success & apply rate \\
\midrule
no governor                 & $0.31$ & --     & $0.09$ & -- \\
governor, $\alpha{=}0$       & $0.88$ & $1.00$ & $0.51$ & $1.00$ \\
cost-blind, $\alpha{=}2$     & $0.19$ & $1.00$ & $0.11$ & $1.00$ \\
cost-aware, $\alpha{=}2$     & $\mathbf{0.40}$ & $0.46$ & $0.10$ & $0.75$ \\
fixed weak, $\alpha{=}2$ (matched) & $0.30$ & $0.45$ & $0.08$ & $0.72$ \\
\bottomrule
\end{tabular}
\end{table}

\section{Demonstrated results II: real-LLM amplification, suppression, and rescue}
\label{sec:gemini}

We ran the induced-cost sweep with Gemini~2.5~Flash as the agent: a cost-blind
governor and cost-aware governors at backoff sensitivity $k\in\{1.5,3.0,6.0\}$,
induced-cost scales $\alpha\in\{0,1,2\}$, six wrong-path tasks designed to engage the governor---a
confident wrong-path hint paired with a short expected answer---and
eight repeats per cell, for $\poweredEpisodes$ episodes with per-episode seeds
mixed with an avalanche finalizer for the induced-cost and backoff random
streams (free-tier limits handled by checkpointed resumption and backoff on
transient errors). Actions were selected through direct calls to Gemini. All
six tasks are included in the pooled results, giving $n=48$ episodes per
condition. One task, C003, never satisfied the substring scoring rule; the
other five had high baseline success. Table~\ref{tab:gemini-tasks} shows all
six tasks individually. An initial three-repeat sweep informed task selection
and exposed a seeding defect affecting the induced-cost draws; the reported
run uses the revised seeds and eight repeats. The tasks therefore form a
targeted stress test rather than a randomly sampled benchmark. The results in
Table~\ref{tab:gemini} show collapse, suppression, and rescue within this
setting.

\paragraph{Amplification.} Under the cost-blind governor, success collapses
monotonically with $\alpha$, $0.83\to0.58\to0.33$, while mean hard blocks rise
$0.1\to4.0\to7.6$ per episode and the answered rate falls to $0.35$: on the
three tasks that storm, blind reaches ${\sim}13$ hard blocks per episode and
most runs never produce an answer. The imposed tightening cost can therefore initiate sustained blocking.

\paragraph{Suppression.} Mean intensity $\bar g$ falls with $\alpha$ under the
cost-aware governor but remains high under the cost-blind one
($0.66\to0.71\to0.74$). At each nonzero tested cost scale, stronger backoff
is associated with lower mean applied intensity: the $\bar g$ drop from
$\alpha{=}0$ to $\alpha{=}2$ is $0.25,0.43,0.49$ for $k=1.5,3.0,6.0$
(Table~\ref{tab:gemini}).

\paragraph{Rescue across backoff settings.} At $\alpha=2$ the cost-blind
governor sits at $0.33$ success; every cost-aware mode beats it, and the rescue
has its highest observed mean at an \emph{intermediate} backoff strength: $0.50,0.75,0.71$ for
$k=1.5,3.0,6.0$, so $k{=}3$ gives the largest rescue ($+0.42$ over blind). The $k{=}1.5$ setting leaves more hard blocks than the stronger settings.
The difference between $k{=}3$ and $k{=}6$ corresponds to two successful
episodes out of 48 per condition, so these data do not establish a unique
optimum or its explanation. The three tasks with sharply reduced success
under blind (T002, E003, C001; $0$--$12.5\%$ success, approximately 11--13 hard
blocks each) all recover to $87.5$--$100\%$ success under $k{=}3$, with fewer
than one hard block per episode on average
(Table~\ref{tab:gemini-tasks}). The aggregate improvement is not uniform:
E002 rises from $7/8$ to $8/8$ successes, whereas C002 falls from $8/8$ to
$5/8$ despite fewer blocks. This task-specific deterioration is a tradeoff in
the observed results; eight repeats do not establish its cause. C003 remains
at zero substring-scored successes in both conditions, although its mean hard
blocks fall from $5.375$ to $0.250$ and its answered count rises from $1/8$ to
$8/8$. This demonstrates reduced blocking and increased answering on C003,
not established answer correctness. The results are
consistent with rescue through renewed opportunities for execution rather
than simple avoidance of induced failures: cost-aware runs can incur more
induced failures because they continue acting.

\paragraph{Counts and interpretation.}
At $\alpha=2$, the success counts are $16/48$, $24/48$, $36/48$, and $34/48$
for blind and $k=1.5,3.0,6.0$, respectively. Each condition contains eight
repeats of each of six fixed tasks, not 48 independently sampled tasks.
The aggregate rates describe this task set; repeated-run uncertainty and
uncertainty about performance on new tasks are distinct. No confidence
intervals or formal tests of the differences between backoff settings are
reported here.

\begin{table}[t]
\centering
\caption{Real-LLM results (Gemini, $\poweredEpisodes$ episodes,
$n=48$ per cell, pooled over all six tasks). Cost-blind amplifies
(success falls, blocks storm, intensity remains high); cost-aware suppresses its
intensity in a monotone $k$ dose-response and rescues success, with the highest observed
success at $k{=}3$.}
\label{tab:gemini}
\begin{tabular}{llcccc}
\toprule
 & $\alpha$ & blind & $k{=}1.5$ & $k{=}3.0$ & $k{=}6.0$ \\
\midrule
\multirow{3}{*}{success}
 & $0$ & $0.83$ & $0.83$ & $0.83$ & $0.83$ \\
 & $1$ & $0.58$ & $0.75$ & $0.77$ & $0.77$ \\
 & $2$ & $0.33$ & $0.50$ & $\mathbf{0.75}$ & $0.71$ \\
\midrule
\multirow{3}{*}{intensity $\bar g$}
 & $0$ & $0.66$ & $0.69$ & $0.69$ & $0.68$ \\
 & $1$ & $0.71$ & $0.62$ & $0.47$ & $0.40$ \\
 & $2$ & $0.74$ & $0.44$ & $0.26$ & $0.19$ \\
\midrule
hard blocks & $2$ & $7.6$ & $2.4$ & $0.3$ & $0.1$ \\
\bottomrule
\end{tabular}
\end{table}

\begin{table}[t]
\centering
\caption{Per-task success and hard blocks at $\alpha=2$ (Gemini, blind vs
$k{=}3$ cost-aware, $n=8$ per cell). All six tasks are shown; success is
reported as a count out of eight to avoid rounding ambiguity. T002, E003,
and C001 recover, whereas C002 has fewer successes under backoff.
Block means are shown to three decimal places, including C003.}
\label{tab:gemini-tasks}
\begin{tabular}{lcccc}
\toprule
 & \multicolumn{2}{c}{successes / 8} & \multicolumn{2}{c}{hard blocks} \\
\cmidrule(lr){2-3}\cmidrule(lr){4-5}
task & blind & $k{=}3$ & blind & $k{=}3$ \\
\midrule
T002 & $0/8$ & $8/8$ & $12.750$ & $0.250$ \\
E003 & $0/8$ & $7/8$ & $12.750$ & $0.750$ \\
C001 & $1/8$ & $8/8$ & $11.375$ & $0.375$ \\
E002 & $7/8$ & $8/8$ & $1.750$ & $0.000$ \\
C002 & $8/8$ & $5/8$ & $1.625$ & $0.125$ \\
C003 & $0/8$ & $0/8$ & $5.375$ & $0.250$ \\
\bottomrule
\end{tabular}
\end{table}

\section{Discussion}

The results show that intervention cost and retained failure history can
jointly determine whether a governor supports or prevents recovery. In the
stochastic baseline comparison, a governor with imposed intervention cost can
perform worse than no governor at all (Table~\ref{tab:baselines}). Adaptive
backoff restores opportunities for execution and improves success for the
persistent policy. Its advantage over an approximately rate-matched fixed
schedule indicates that average intervention frequency alone is insufficient
to explain the result in that setting.

The Gemini experiment extends the collapse-and-rescue observation to a learned
action policy using the same governor and cost construction. It does not
include a fixed-weak or ungoverned arm, so the matched-rate adaptive comparison
remains a stochastic-substrate result. The agreement between two cost
implementations also concerns their shared governor-visible failure
construction, rather than independently measured intervention costs.

The connection to IRAM-$\Omega$-Q remains motivational and structural. These
experiments do not validate its architecture or reproduce its numerics. The
practical finding is that permanent blocking can preserve the effect of a
temporary fault, and that intervention scheduling changes whether an agent can
escape that state. Error-aware alternatives, including bounded retries for
transient failures, are important comparisons for determining when adaptive
backoff is the appropriate remedy.

\section{Limitations}

The intervention cost is experimentally imposed, and the backoff rule has
direct access to the induced-event label. Whether comparable costs arise
naturally, and how a governor could distinguish them from ordinary tool
failures, remain open questions. The stochastic agent is hand-coded; the real
LLM results use one model, Gemini~2.5~Flash, and six closely related tasks
selected or designed using pilot experience. Eight repeats per condition and
task quantify variation in this setting but do not establish broad task or
model generality. The reported means are descriptive, and the best observed
backoff setting should not be interpreted as a statistically established
optimum.

The one-sided quadratic cost law and intensity ladder are modeling choices,
not calibrated physical or computational costs. Other cost shapes and timing
rules may yield different behavior. The benchmark governor also knows the
expected target path when judging recovery. Correctness is scored by a
normalized substring match against the expected answer, which can reject
correct paraphrases or differently expressed units. The persistent zero score
on C003 does not by itself distinguish a scoring mismatch from an incorrect
answer. Without complete model responses, the checkpoint alone cannot resolve
that distinction. Finally, the
fixed-weak comparison is limited to the stochastic substrate and a particular
application probability. Broader comparisons with bounded retry, expiring
failure memory, and alternative fixed schedules would clarify the relative
benefits of cost-aware backoff.

\section{Conclusion}

A supervisory governor can sustain task failure when an intervention-induced
tool error is retained as a reason to block subsequent attempts. In the
controlled environment studied here, this interaction produces block storms
and lower completion rates. Feedback-conditioned backoff reduces blocking and
can restore success on both a stochastic-policy agent and a Gemini agent.
On the stochastic substrate it also outperforms an approximately rate-matched
fixed weak governor; on Gemini the highest observed success occurs at an
intermediate tested backoff strength. These findings motivate evaluating
governors together with their intervention costs, error memory, and recovery
opportunities.

\section*{Data and code availability}
The ancillary files accompanying this preprint contain the Gemini episode
checkpoint and the stochastic aggregate summaries used for the reported
tables, together with a script that regenerates their numerical summaries.
The governed-agent framework, induced-cost agents, experiment harnesses, and
analysis scripts are maintained in the CASAgentGovernor repository at
\url{https://github.com/CriticalAttentionSystems/CASAgentGovernor}.
The Gemini checkpoint records episode-level metrics; it does not contain the
complete model responses or tool trajectories.


\begin{thebibliography}{9}
\small
\setlength{\itemsep}{3pt}

\bibitem{astrom2008feedback}
K.~J. \r{A}str\"om and R.~M. Murray.
\newblock \emph{Feedback Systems: An Introduction for Scientists and Engineers}.
\newblock Princeton University Press, 2008.

\bibitem{ziegler2026iramomegaq}
V.~Ziegler.
\newblock \emph{IRAM-$\Omega$-Q: A Computational Framework for Uncertainty
Regulation in Adaptive Agents}.
\newblock arXiv:2603.16020, 2026.

\bibitem{ziegler2026carryover}
V.~Ziegler.
\newblock \emph{When Regulation Has Memory: Hysteresis and Control Burden
in Artificial Agency}.
\newblock arXiv:2606.30975, 2026.

\bibitem{ziegler2026switching}
V.~Ziegler.
\newblock \emph{Intermittent Control Is Not Diluted Control: A Switching Effect
in Artificial Agency}.
\newblock arXiv:2607.17432, 2026.

\bibitem{li2026autonomytax}
S.~Li and Y.~Zhao.
\newblock \emph{The Autonomy Tax: Defense Training Breaks LLM Agents}.
\newblock arXiv:2603.19423, 2026.

\bibitem{kim2026causalarmor}
M.~Kim et al.
\newblock \emph{CausalArmor: Efficient Indirect Prompt Injection Guardrails
via Causal Attribution}.
\newblock arXiv:2602.07918, 2026.

\bibitem{pizarro2023multistep}
F.~Pizarro Bejarano, L.~Brunke, and A.~P.~Schoellig.
\newblock \emph{Multi-Step Model Predictive Safety Filters: Reducing Chattering
by Increasing the Prediction Horizon}.
\newblock arXiv:2309.11453, 2023.

\bibitem{sutton2018rl}
R.~S. Sutton and A.~G. Barto.
\newblock \emph{Reinforcement Learning: An Introduction}.
\newblock MIT Press, second edition, 2018.

\end{thebibliography}
\end{document}